\documentclass[sigconf]{acmart}

\AtBeginDocument{%
  }

\setcopyright{acmcopyright}
\copyrightyear{2023}
\acmYear{2023}
\acmDOI{https://doi.org/10.1145/3543873.3587630}

\acmConference[Workshop on Interactive and Scalable Information Retrieval Methods for eCommerce]{The Web Conference 2023}{Apr 30--May 04,
  2023}{Austin, TX}

\title{Universal Model in Online Customer Service}

\author{Shu-Ting Pi}
\email{shutingp@amazon.com}
\affiliation{%
  \institution{Amazon}
  \city{Cupertino}
  \state{CA}
  \country{USA}
  \postcode{95014}}

\author{Cheng-Ping Hsieh}
\authornote{Work was completed while CP was an intern at Amazon}
\email{chengpi@amazon.com}
\affiliation{%
  \institution{Amazon}
  \city{Seattle}
  \state{CA}
  \country{USA}
  \postcode{98109}}

\author{Qun Liu}
\email{qunliu@amazon.com}
\affiliation{%
  \institution{Amazon}
  \city{Seattle}
  \state{CA}
  \postcode{98109}
  \country{USA}
  }

\author{Yuying Zhu}
\email{imyuying@amazon.com}
\affiliation{%
  \institution{Amazon}
  \city{Seattle}
  \state{CA}
  \country{USA}
  \postcode{98109}}

\begin{document}

\renewcommand{\shortauthors}{Pi et al.}

\begin{abstract}
Building machine learning models can be a time-consuming process that often takes several months to implement in typical business scenarios. To ensure consistent model performance and account for variations in data distribution, regular retraining is necessary. This paper introduces a solution for improving online customer service in e-commerce by presenting a universal model for predicting labels based on customer questions, without requiring training. Our novel approach involves using machine learning techniques to tag customer questions in transcripts and create a repository of questions and corresponding labels. When a customer requests assistance, an information retrieval model searches the repository for similar questions, and statistical analysis is used to predict the corresponding label. By eliminating the need for individual model training and maintenance, our approach reduces both the model development cycle and costs. The repository only requires periodic updating to maintain accuracy.

\end{abstract}
\maketitle
\section{Introduction}
E-commerce websites handle a vast number of online customer service requests daily. In a typical online customer service, customers initially interact with a chatbot by asking the customers some questions that identifies their intent. Typically, intent is classified based on product or service. For example, for a travel solution web service, their chatbot could attempt to classify requests as rental car, airline, hotel, or cruise issues. The chatbot would then route the customer to an agent specialized in the requested product to provide assistance. The process of interaction between customers and agents is also similar across different companies. Agents begin with greetings and request details about the customer's questions. They then engage in diagnosis and finally conclude with some closing words.

Generally, each request takes several minutes, making it the most time-consuming part of the e-commerce business. Therefore, improving performance of machine learning models in the early stage before the customer reaches an agent can help customers to save considerable time and better use the bandwidth of available agents. However, training a machine learning model can be a lengthy process. It could take several months to work and the model has to be retrained regularly to adapt new data distributions. It becomes a severe issue when emerging events occur such as new products or new services launch. Therefore, we propose a flexible, universal model that can predict any label (categorical or continuous) based on customer input text, without the need for training. This model is designed to be quickly implemented in situations where a fast onboarding process is necessary.

Our universal model is based on the idea that customer service agents' reactions to a customer contact are mainly influenced by the issues they're dealing with. This means that two contacts with similar topics should have similar probabilities for outcomes, such as handling time, need for third-party assistance, or need for transfer, regardless of the specific agents and customers involved. To put this hypothesis into practice, we've built a repository that collects millions of customer questions and their corresponding attributes or labels from different customer service contacts. When a new customer contact comes in, their input question is used to search for the most similar questions in the repository. The corresponding attributes or labels are then retrieved and statistical measures, such as the mode, mean, or median, are computed to make predictions.

The universal model's workflow, depicted in Fig. 1, consists of two phases: 1) creating a repository of customer inquiries and their characteristics, and 2) constructing a retrieval-based model to identify comparable questions in the repository and generate predictions. To implement this concept, a machine-learning model that can tag the primary inquiry in a customer's transcript is a crucial component of the article.

The article is structured as follows: Section 2 introduces the development of a sentence tagging model to automatically identify the main question in a contact transcript. In Section 3, we describe the construction of our universal model using the question repository created by the sentence tagging model. We also compare the performance between our universal model and traditional supervised learning models. Finally, in Section 4, we draw conclusions.

\begin{figure*}
  \centering
  \includegraphics[width=0.9\textwidth]{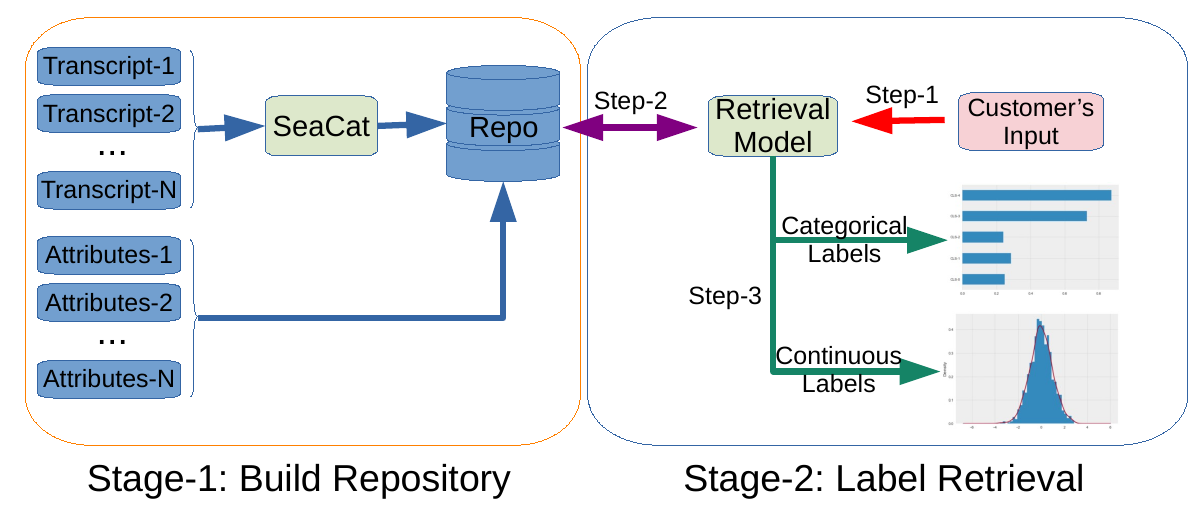}
  \caption{The Universal Model workflow has two stages. In Stage 1, customer questions are extracted from transcripts using the SeaCat Model, and their attributes are combined to form a repository. In Stage 2, the process begins with the customer entering their question or request into a retrieval model. The retrieval model then searches for the Top-N most similar questions and retrieves their attributes from the repository. Next, the distributions of the retrieved attributes are analyzed to make predictions. Categorical labels are predicted using normalized frequency (probability), while continuous labels are predicted using their median.}\label{workflow}
\end{figure*}  

\textbf{Our Contributions} The key contributions of our work include 1) the Development of a sentence tagging model that automatically extracts customers' main questions from transcripts, without the need for additional human annotation. This is a pioneering approach in the field. 2) Creation of a comprehensive repository of customer questions, providing insight into customer needs. 3) Proposal of a universal model, a retrieval-based approach that can predict new labels without retraining the model.       

\textbf{Related Works} Our work is related to several areas in the literature, including sentence tagging with weakly supervised/unsupervised learning, information retrieval, and multi-task learning in text data. One similar study is the dialogue QA matching by Jia et al. \citep{dialoque}, which uses a deep learning approach called mutual attention to address a similar problem. Our work is also inspired by unsupervised QA harvesting \citep{QA_refining} and the dynamical chunk reader \citep{chunk_reader}. The latter uses an unsupervised approach to extract question-answer pairs from Wikipedia and apply them to the SQUID dataset \citep{SQUAD1, SQUAD2}. In terms of information retrieval and multi-task learning in text data, Liu et al \citep{multidomain} combine deep learning, multi-task learning, and multi-domain query classification, which is related to our approach. Furthermore, Wang et al \citep{rethinking} used retrieval, rethinking, and multi-task learning as a framework to solve QA problems, providing another important reference for our work.

\section{Tagging Customer's Questions} 

\subsection{The Dataset}
To date, there has been no publicly available dataset related to customer service transcripts. As such, we partnered with Amazon's customer service team to initiate this research, utilizing a dataset from the MessageUS channel, an online chat platform that enables customers to communicate with Amazon's customer service agents. We eliminated all Amazon-specific features and only retained three features that any online customer service platform must have: the product/service, the handling time, and the customer-agent transcripts.

It is worth noting that the dataset only contains customer text data, with all confidential information, such as names and account details, anonymized to protect privacy before being shared with researchers. While the dataset is from Amazon's database, its format is general, making the methodology presented in this article applicable to other use cases as well.

\begin{figure*}
  \centering
  \includegraphics[width=0.8\textwidth]{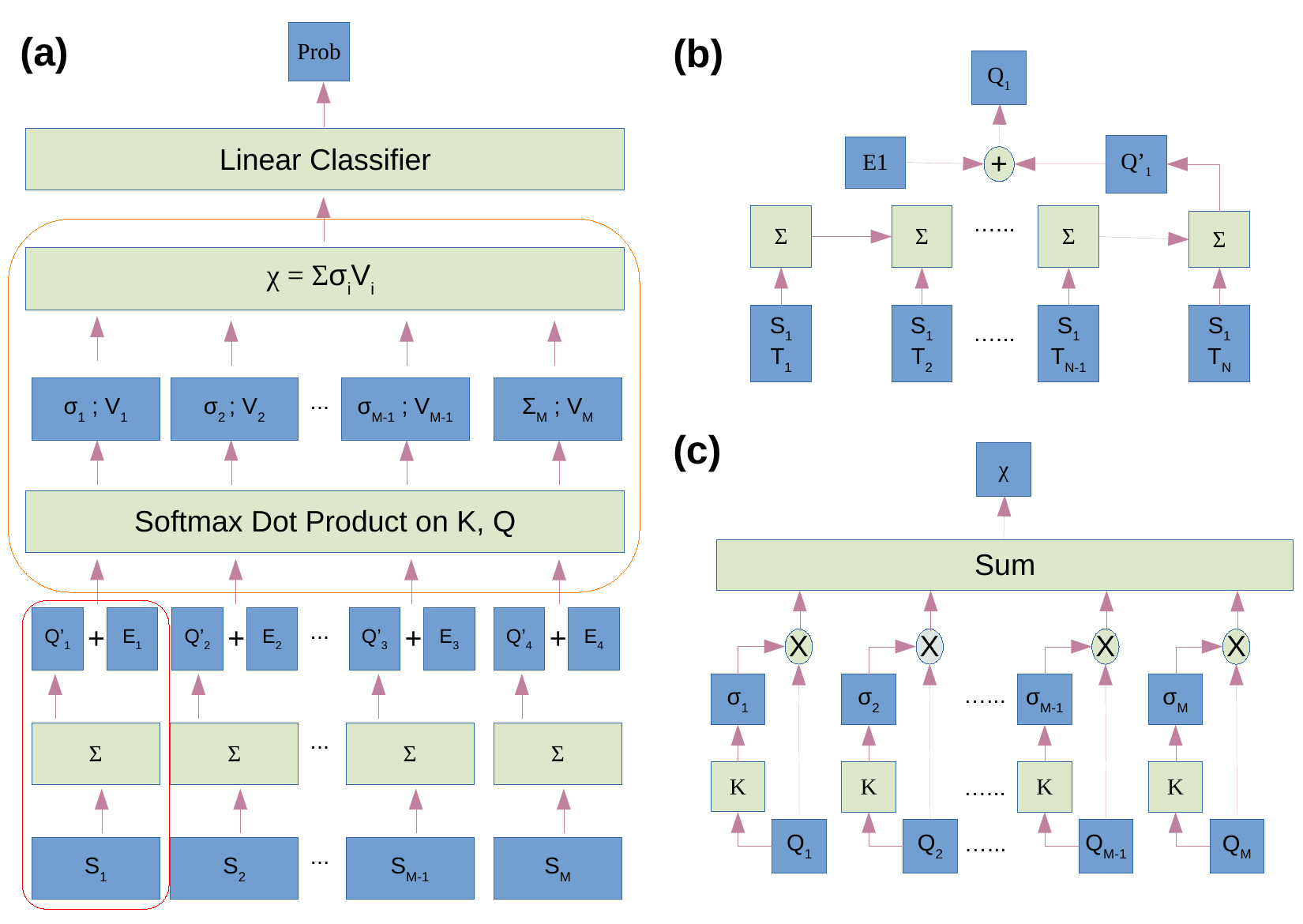}
  \caption{The SeaCat Model. The model consists of blue blocks, representing tensors, and green blocks representing operators. (a) The neural network is comprised of sentence tensors, $S_{i}$, and a sequence model, $\Sigma$, which outputs $Q^{'}{i}$. The position embedding vector $E{i}$ is combined with $Q^{'}{i}$ to create the sentence embeddings $Q{i}$. Finally, a linear classifier predicts the product/service. (b) The red block in (a) is described in detail. Tensor notation ($S_1;T_n$) refers to the $n$-th token in sentence $S_1$. The sequence model $\Sigma$ processes each word in a sentence using a time-distributed wrapper to handle multiple sentences. (c) The orange block in (a) is explained. A dense layer, $K$, with a softmax activation function is applied to all sentence embedding vectors $Q_{i}$ (via a time-distributed wrapper) to calculate attention scores $\sigma_{i}$. Note that $Q_{i}$ is equivalent to $V_{i}$.}\label{SeaCat}
\end{figure*}

\subsection{Sentence Attention in Dialogue Transcripts}
Our hypothesis for the sentence tagging model is that the customer's main question sentence can be identified based on two key features: (1) it typically appears in the first few sentences of the customer's interaction with the agent, and (2) it contains the most relevant information about the topic of the interaction. This hypothesis can be transformed into a well-defined machine learning problem: \textbf{to identify which customer sentences located near the agent's first response are most crucial for a machine learning classifier to accurately predict the product/service of the interaction, a general label that applies to any online customer service}.

Suppose we have a dataset of conversational transcripts, where each transcript is labeled with its related topic. Our goal is to train a deep learning model that can accurately predict the label while also extracting the importance of each sentence in the transcript. To achieve this, we propose a machine-learning model called SeaCat (Sentence Attention of Customer-Agent Transcript). Unlike traditional text classification models, which embed an article as a 2D tensor of size $\mathbb{N}{at} \times \mathbb{N}{we}$, where $\mathbb{N}{at}$ is the number of tokens in the article and $\mathbb{N}{we}$ is the dimension of the word embedding, SeaCat embeds each article as a 3D tensor of size $\mathbb{N}{as} \times \mathbb{N}{st} \times \mathbb{N}{we}$, where $\mathbb{N}{as}$ is the number of sentences in the article and $\mathbb{N}_{st}$ is the number of tokens per sentence. To make the inputs fixed-size, we apply zero padding to the tensors, even though padding is optional in deep learning frameworks. This makes the problem more well-defined.

Next, we extract the sentence embedding vectors of each sentence using a sequence model such as BERT \citep{bert} or LSTM \citep{lstm}. This is accomplished by treating each sentence as a temporal slice and applying a time-distributed wrapper (TensorFlow v2.8) to the sequence model, ensuring it only receives one sentence per time step. The resulting tensor has shape $\mathbb{N}{as} \times \mathbb{N}{se}$, where $\mathbb{N}{as}$ represents the number of sentences per article and $\mathbb{N}{se}$ represents the size of the sentence embedding in the sequence model. The meaning of this tensor is straightforward, each sentence is embedded as a vector with $\mathbb{N}{se}$ dimensions, and there are $\mathbb{N}{as}$ sentences per article.

To determine the importance of each sentence, we define the sentence attention as below \citep{transformer}. 
\begin{equation}
 Attention(\overrightarrow{K},\widehat{Q},\widehat{V)} = \overrightarrow{\chi} = softmax( \frac{\widehat{Q} \cdot \overrightarrow{K}}{\sqrt{d_{k}}})^{T} \cdot \widehat{V} = \overrightarrow{\sigma}\cdot \widehat{V}     
\end{equation}

, where $\widehat{Q}$ is a query tensor, $\overrightarrow{K}$ is a key vector and  $\widehat{V}$ is a value tensor. In our problem, we define $\widehat{Q}$ = $\widehat{V}$ = the sentence embeddings tensors (in general, one can train another tensor $\widehat{V}$ as the value tensor but there is no significant difference on model performance), i.e. the output tensor of the sequence model $\widehat{Q}^{'}$ plus sentence position embedding tensor $\widehat{E}$ (see next section). $\overrightarrow{K}$ is the key vector that takes the inner product with the output of the sequence model. If so, the shape of $\widehat{Q}$ and $\widehat{V}$ are the same with the sentence embedding, i.e. $\mathbb{N}_{as}\times\mathbb{N}_{se}$. As for $\overrightarrow{K}$, it is a vector with dimension $\mathbb{N}_{se}$. As a result, the attention value $\overrightarrow{\sigma}\cdot \widehat{V}$ is a vector with size $\mathbb{N}_{se}$ and the sum of its elements must equal 1, due to the application of the softmax function. It's worth noting that this definition differs slightly from the scalar dot-product attention used in the Transformer \citep{transformer}, as the Key $\overrightarrow{K}$ is a vector rather than a matrix. This change was made because our objective is to determine sentence weight, not self-attention, so attention weights between sentences are not taken into account.

The meaning of the attention vector $\overrightarrow{\chi}$ is straightforward. As shown in Fig. 2, each sentence in a transcript is embedded as a vector $\overrightarrow{V}_{i}$ (the $i$-th row of the 2D tensor $\widehat{V}$), and $\overrightarrow{\chi}$ is the linear combination of all sentences, calculated as $\overrightarrow{\chi} = \sum_{i}\sigma_{i} \overrightarrow{V}_{i}$ (with the constraint that $\sum_{i}\sigma_{i}=1$). The attention vector $\overrightarrow{\chi}$ is then fed into a fully connected layer with a softmax activation function to predict the product/service of the transcript. In other words, the attention weight $\sigma_{i}$ indicates the importance of the $i$-th sentence in determining the product/service, with higher weights assigned to sentences that contain more critical information. 

\subsection{Sentence Position Embedding}
The "bag of sentences" model generates the attention weight $\overrightarrow{\sigma}$ through a time-distributed dense layer $\overrightarrow{K}$, which does not consider the position of sentences. However, based on our observation, we found that most customer questions appear in the early sentences during their interaction with agents. This indicates that sentence positions can have an impact on attention weights. To capture this information, we aim to incorporate sentence position information into the model.

To include sentence position information in the model, there are two options: 1) using a sequence layer like LSTM instead of a time-distributed dense layer for $\overrightarrow{K}$, which automatically incorporates sentence positions, or 2) introducing position embedding vectors to provide additional information about sentence positions. After experimentation, option 2 was found to be faster and more effective, so we will focus on this approach.

To generate position embedding tensors, we first assign an index to each sentence, ranging from -$\mathbb{N}_{as}$ to +$\mathbb{N}_{as}$, representing the number of sentences between the current sentence and the first sentence spoken by the agent. For example, a sentence that is five steps before the agent's first sentence would have an index of -5, while a sentence that is five steps after the agent's first sentence would have an index of +5. To avoid negative indices, we then shift the indices by $\mathbb{N}{as}$, resulting in an allowed index range of 0 to +$2\mathbb{N}{as}$, with the sentence that has an index of $\mathbb{N}_{as}$ being the agent's first sentence.

The sentence index assigned to each sentence is used to generate the position embedding tensors. Using the concept of word position embedding from BERT \citep{bert}, a mathematical formula is used to embed the sentence positions. For the $i$-th sentence, its position embedding vector $E_{i}$ is computed using the formula, with the $p$-th component represented by $E_{i}(2p)$ and $E_{i}(2p+1)$:
\begin{align}
E_i(2p) &= sin(i/10000^{2p/d_{pos}})\\    
E_i(2p+1) &= cos(i/10000^{2p/d_{pos}})
\end{align}
The dimension of the embedding vector, $d_{pos}$, is also defined. The position embedding vector is then added to the sequence model's output $Q^{'}_{i}$ to form the sentence embedding $Q_{i}$, thereby incorporating information about sentence positions into the calculation of attention weights.

\subsection{Experiments}
We have created a dataset of 500,000 contact transcripts to evaluate the performance of our sentence tagging model. The product/service, consisting of 152 classes, is used as the labels for the transcripts. Each transcript was padded with zeros to consist of 64 sentences, each with 128 words, resulting in $\mathbb{N}_{as} = 64$ and $\mathbb{N}_{st}=128$. Sentence embedding vectors were generated by treating each sentence as a temporal slice and using the pre-trained DistilBERT model \citep{distill_bert} as the sequence model, represented as $\Sigma$ in Fig. 2. The sentence position embedding vectors were generated based on the padded transcripts, with the sentence position index ranging from 0 to 128, where index 64 represents the agent's first sentence, and with a dimension of $d_{pos}=768$, equal to the output dimension of DistilBERT.

After completing the training process, we can determine the attention weights $\sigma_{i}$ for all customer sentences in each transcript. To identify the customer's main question, we proposed two methods:

\begin{itemize}
    \item Select the customer sentence with the highest attention weight in the transcript.
    \item Pick the customer sentence with the highest attention weight that is two steps before or after the agent's first sentence (Note: Based on our experiment, $\pm2$ yielded the best result. However, in different business scenarios, adjustments to this number may be necessary).
\end{itemize}

We evaluated the performance of these methods by testing the models on a human-annotated dataset of 4000 transcripts. The accuracy of the first method was 76.68\% (i.e., the main question was correctly tagged in 3372 transcripts), while the second method had an accuracy of 84.3\%. We also compared the performance of the SeaCat model without sentence position embedding, which had an accuracy of 83.4\%, slightly lower than the previous result.

An example of sentence tagging in a transcript is provided in the appendix. In order to safeguard customer privacy, all of the sentences and product/service names demoed in this article are derived from a synthetic research dataset designed solely for demonstration purposes, and do not represent any actual customer service data collected from Amazon. 

In the figure, the customer sentences are assigned attention weights between 0-1, and the sentence with the highest score near the agent's first sentence is considered the main question. It is worth noting that sentences 2 and 9 are similar and have similar scores, but the difference between them increases when considering sentence position (0.20 vs 0.22 to 0.17 vs 0.25). This may be a result of the position embedding's influence on the model's focus on sentences closer to the agent's first sentence.

The attention weight not only helps identify the customer's main question sentence, but also makes our classifier more interpretable. The high accuracy supports our hypothesis that the most critical sentences in determining the topic are often the customer's main question sentences. Given that the performance of rule 2 is significantly better than rule 1, we will use it as our standard model in future discussions.

\begin{figure*}
  \centering
  \includegraphics[width=1.0\textwidth]{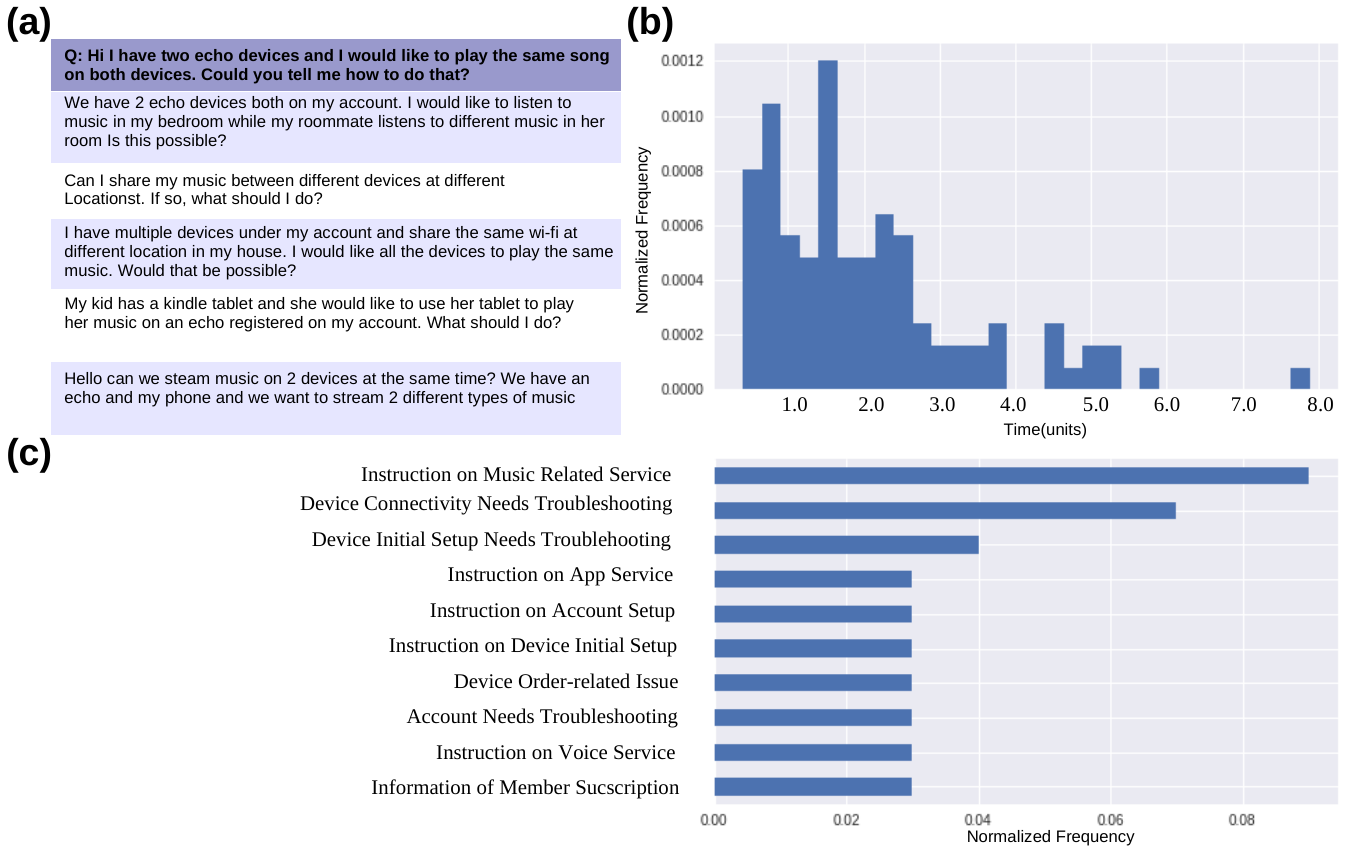}
  \caption{Illustration of how the Universal Model operates. (a) The top-5 most similar questions are retrieved from the 12K repository for a given customer inquiry (as shown in the first row). (b) The distribution of handling times for the top-100 retrieved similar contacts is displayed. The median of this distribution will be used as the predicted handling time. (c) The distribution of product/service for the top-100 retrieved similar contacts is also shown. The product/service with the highest frequency will be the predicted label. Note: All the sentences and the product/service names are derived from a synthetic research dataset for demonstration purposes, and do not represent any actual customer service data collected from Amazon. }\label{example}
\end{figure*}

\section{The Universal Model}
The goal of the universal model is to build a highly flexible machine-learning model that can adapt to new data distribution and predict new labels using customer input questions without any model retraining. In the following, we detail how the universal model works. 
\begin{table*}[t]
\centering
\begin{tabular}{| c | c c c c | c c c|}
\hline
{} & \multicolumn{4}{c|}{Classification (product/service)} & \multicolumn{3}{c|}{Regression (handle time)} \\
 {\textbf{Model}}& Top-1 & Top-5 & Top-10 & Top-15 & 3 unit & 5 unit & 7 unit   \\
\hline
\hline
RF(3K)&22.5& 53.8& 65.3& 72.5& 19.5& 33.3& 47.4\\
RF(6K)&25.4& 55.2& 67.7& 74.7& 20.2& 33.4& 47.5\\
RF(9K)&27.2& 57.8& 69.2& 73.3& 20.7& 34.1& 47.6\\
RF(12K)&27.4& 57.9& 69.7& 77.5& 21.4& 34.4& 47.9\\
\hline
\hline
UM(3K)&24.1& 55.95& 70.1& 77.15& 21.6& 37.3& 52.5\\
UM(6K)&26.2& 58.4& 72.85& 78.5& 22.2& 37.7& 52.8\\
UM(9K)&27.9& 59.8& 73.3& 81.9& 22.6& 37.9& 53.4\\
UM(12K)&28.1& 61.25& 74.1& 82.9& 22.8& 38.3& 53.4\\
\hline
\hline
\end{tabular}
\caption{Performance of Random Forest (RF) and the universal model (UM) using 3K, 6K, 9K, and 12K data, respectively, to predict the product/service and handle time. Prediction on product/service is evaluated via the Top-1, Top-5, Top-10, and Top-15 accuracy (among 152 classes, in a percentage unit). Prediction on handle time is evaluated via the percentage of the contacts where the absolute error is less than 3, 5, and 7 units, the metrics that are frequently used for online customer service. }
\label{tab:2}
\end{table*}

\subsection{Build the Question repository}
As mentioned in the previous section, the dataset only contains three features of customer contacts: the conversation transcript, handle time, and product/service of customer's issue. We removed all Amazon-specific features to ensure our ideas apply to all other online customer service platforms. Using our sentence tagging model, which identifies the customer's main question, we constructed a database with thousands or millions of customer questions and their corresponding labels, such as the product/service and handle time we adopted here. This database serves as the backbone of our universal model. When a new customer inputs their question, we can use the database to retrieve the most similar questions and make predictions based on the retrieved labels.

\subsection{Label Retrieval}
Once the repository is established, the customer's question sentence serves as the "key." Upon entering the online chat customer service system, the customer's input question serves as the "query." This query is then fed to the retrieval model, $\mathcal{F}$, to calculate the similarity between the query and each question sentence in the repository. For example, when a customer inputs "my online order was not delivered on time" the query is compared to each question in the repository to find the most similar match.

Based on the retrieved similar questions, their corresponding labels are also retrieved to make predictions. For instance, the top 100 similar questions and their labels could be analyzed with statistical methods. In the case of categorical labels, the predicted result is the category with the highest frequency or probability. For continuous labels, the median is used as the predicted result. This is a concise explanation of how the universal model operates.

The universal model can be built using various unsupervised options for the retrieval model, such as the Okapi BM25 model \citep{bm25_1,bm25_2}. This makes the creation of the model fully unsupervised, as the repository still requires labels, but eliminates the need for frequent retraining to adapt to new data distributions or predict new labels. All that is required is an update to the repository with the latest data, making the universal model highly flexible and able to quickly adapt to new business scenarios without the need for supervised model retraining.

\subsection{Experimental Results}
To determine the similarity between a customer's input and the questions stored in our repository, we utilize the Okapi BM25 model as our retrieval engine \citep{bm25_1,bm25_2}. Despite the fact that deep learning retrieval models show slightly better performance on the metrics we care about, we opt for a traditional machine learning model to achieve more interpretable results. Given an input question sentence $q$ (referred to as the query) and a question sentence $k$ (referred to as the key) in the repository, the similarity score is calculated as follows:

\begin{align}
\mathcal{F}(q,k) = \sum_{i=1}^{N} IDF(q_{i})\frac{TF(q_i|k)(\alpha+1)}{TF(q_i|k)+\alpha\times (1-\beta+\beta\times \frac{|k|}{\langle |k| \rangle})}    
\end{align}

, where $TF(q_i|k)$ is the term frequency of $i$-th word token $q_i$ in $q$ found in $k$, $IDF(q_{i})$ is the inverse document frequency of token $q_i$ in the whole repository \citep{tfidf_1,tfidf_2}, $|k|$ is the number of word tokens in $k$, and $\langle |k| \rangle$ is the average length of each document in the repository. $\alpha$ and $\beta$ are free parameters that must be tuned based on your data. Here we use $\alpha=1.2$ and $\beta=0.75$.   

To assess the performance of our model, we employ the SeaCat model to gather 3K, 6K, and 12K questions. We use these question repositories with varying sizes to build the universal model and predict the product/service (a categorical label) and handle time (a continuous label) based on the Top-100 most similar questions retrieved from the repository and their corresponding labels. As shown in Fig.3, the universal model operates as follows. In Fig.3(a), the Top-5 most similar questions for a given query are depicted (based on the 12K repository). The retrieved questions are very similar to the input query. In Fig.3(b) and (c), the distributions of the labels on the Top-100 similar questions are displayed. These distributions serve as the predictions of the universal model and reflect the labels and their distributions.

To compare the performance of our model, we also train Random Forest classifier and Random Forest regressor using the question sentences in the repository as inputs and their corresponding labels as outputs \citep{random_forest_1,random_forest_2}. These models serve as benchmarks for our tasks. To evaluate the performance of the classifier in predicting product/service (categorical label), we use the Top-N accuracy metric, which measures the accuracy of the top N predictions (152 classes in total). For the regressor in predicting handling time (continuous label), we use a metric that is more relevant to online customer service scenario, which is the percentage of contacts where the absolute error is less than 3, 5, and 7 units (Here, we use the term "unit" to conceal the actual time unit for the purpose of safeguarding business confidentiality). The results of these evaluations are presented in Table 1 using a test set of 4000 examples.

Based on the results, we found that both the Random Forest and the Universal model show improved performance with increasing data. Despite being developed through a fully unsupervised approach, our Universal model outperforms the Random Forest in both tasks, demonstrating the potential of correctly designed information retrieval as a classifier or regressor. The universal model is essentially an alternative form of K-Nearest Neighbor (KNN) method, using the BM25 model to calculate the distance between keys and queries, with the number of selected similar questions serving as the hyper-parameter "K" in KNN \citep{knn_1, knn_2}.

\section{Conclusion}
In conclusion, we have presented a unique machine learning approach for extracting customer questions from transcripts and using them to create a universal model for predicting categorical or continuous labels in a repository. The results indicate that the unsupervised universal model outperforms conventional supervised models such as Random Forest. The model also demonstrated improvement when the size of the repository was increased. This paves the way for building a versatile machine learning platform that can support various business applications.

Our work also has a scientific contribution by demonstrating the application of sentence-level attention mechanism in conversational transcripts, which is an understudied area. The token-level attention mechanism was successfully extended to the sentence level, and it proved to be effective in resolving a real-world problem. In the future, we aim to explore the potential of self-attention mechanism at the sentence level in various situations.

\bibliographystyle{apalike}
\bibliography{citation.bib}

\begin{thebibliography}{}

\bibitem[Altman, 1992]{knn_2}
Altman, N.~S. (1992).
\newblock An introduction to kernel and nearest-neighbor nonparametric
  regression.
\newblock {\em The American Statistician}, 46(3):175--185.

\bibitem[Beel et~al., 2016]{tfidf_2}
Beel, J., Gipp, B., Langer, S., and Breitinger, C. (2016).
\newblock Research-paper recommender systems: a literature survey.
\newblock {\em International Journal on Digital Libraries}, 17:305--338.

\bibitem[Devlin et~al., 2017]{bert}
Devlin, J., Chang, M.-W., and Kenton~Lee, K.~T. (2017).
\newblock Bert: Pre-training of deep bidirectional transformers for language
  understanding.
\newblock {\em Advances in neural information processing systems}, 30.

\bibitem[Fix and Hodges, 1951]{knn_1}
Fix, E. and Hodges, J.~L. (1951).
\newblock Discriminatory analysis. nonparametric discrimination: Consistency
  properties.
\newblock {\em USAF School of Aviation Medicine, Randolph Field, Texas}.

\bibitem[Ho, 1995]{random_forest_1}
Ho, T.~K. (1995).
\newblock Random decision forests.
\newblock {\em Proceedings of 3rd international conference on document analysis
  and recognition}, 1:278--282.

\bibitem[Ho, 1998]{random_forest_2}
Ho, T.~K. (1998).
\newblock The random subspace method for constructing decision forests.
\newblock {\em IEEE transactions on pattern analysis and machine intelligence},
  20(8):832--844.

\bibitem[Hochreiter and Schmidhuber, 1996]{lstm}
Hochreiter, S. and Schmidhuber, J. (1996).
\newblock Lstm can solve hard long time lag problems.
\newblock {\em Advances in neural information processing systems}, 30:473--479.

\bibitem[Jia et~al., 2021]{dialoque}
Jia, Q., Zhang, M., Zhang, S., and Zhu, K.~Q. (2021).
\newblock Matching questions and answers in dialogues from online forums.
\newblock {\em arXiv}, page 2101.03064.

\bibitem[Jones et~al., 2000a]{bm25_1}
Jones, K.~S., S.Walker, and S.E.Robertsonb (2000a).
\newblock A probabilistic model of information retrieval: development and
  comparative experiments: Part 1.
\newblock {\em Information Processing and Management}, 36(6):779--808.

\bibitem[Jones et~al., 2000b]{bm25_2}
Jones, K.~S., S.Walker, and S.E.Robertsonb (2000b).
\newblock A probabilistic model of information retrieval: development and
  comparative experiments: Part 2.
\newblock {\em Information Processing and Management}, 36(6):809--840.

\bibitem[Li et~al., 2020]{QA_refining}
Li, Z., Wang, W., Dong, L., Wei, F., and Xu, K. (2020).
\newblock Harvesting and refining question-answer pairs for unsupervised qa.
\newblock {\em Proceedings of the 58th Annual Meeting of the Association for
  Computational Linguistics}, pages 6719--6728.

\bibitem[Liu et~al., 2021]{multidomain}
Liu, Y., Lee, J., Zhu, L., Ling~Chen, H.~S., and Yang, Y. (2021).
\newblock A multi-mode modulator for multi-domain few-shot classification.
\newblock {\em IEEE/CVF International Conference on Computer Vision (ICCV)},
  pages 8453--8462.

\bibitem[Rajaraman and Ullman, 2012]{tfidf_1}
Rajaraman, A. and Ullman, J.~D. (2012).
\newblock Massive data mining.
\newblock {\em Cambridge University Press}.

\bibitem[Rajpurkar et~al., 2018]{SQUAD2}
Rajpurkar, P., Jia, R., and Liang, P. (2018).
\newblock Know what you don't know: Unanswerable questions for squad.
\newblock {\em arXiv}, page 1806.03822.

\bibitem[Rajpurkar et~al., 2016]{SQUAD1}
Rajpurkar, P., Zhang, J., Lopyrev, K., and Liang, P. (2016).
\newblock Squad: 100,000+ questions for machine comprehension of text.
\newblock {\em arXiv}, page 1606.05250.

\bibitem[Sanh et~al., 2020]{distill_bert}
Sanh, V., Debut, L., and Julien~Chaumond, T.~W. (2020).
\newblock Distilbert, a distilled version of bert: smaller, faster, cheaper and
  lighter.
\newblock {\em arXiv}, (1910):01108.

\bibitem[Vaswani et~al., 2017]{transformer}
Vaswani, A., Shazeer, N., Parmar, N., Uszkoreit, J., Jones, L., Gomez, A.~N.,
  Kaiser, L., and Polosukhin, I. (2017).
\newblock Attention is all you need.
\newblock {\em Advances in neural information processing systems}, 30.

\bibitem[Wang et~al., 2021]{rethinking}
Wang, Z., Ng, P., Nallapati, R., and Xiang, B. (2021).
\newblock Retrieval, re-ranking and multi-task learning for knowledge-base
  question answering.
\newblock {\em Conference of the European Chapter of the Association for
  Computational Linguistics}, pages 347--357.

\bibitem[Yu et~al., 2016]{chunk_reader}
Yu, Y., Zhang, W., Hasan, K., Yu, M., Xiang, B., and Zhou, B. (2016).
\newblock End-to-end answer chunk extraction and ranking for reading
  comprehension.
\newblock {\em arXiv}, page 1610.09996.

\end{thebibliography}

\newpage
\onecolumn

\section{Appendices}
\setcounter{figure}{0}
\begin{figure*}[!htb]
  \centering
  \includegraphics[width=0.9\textwidth]{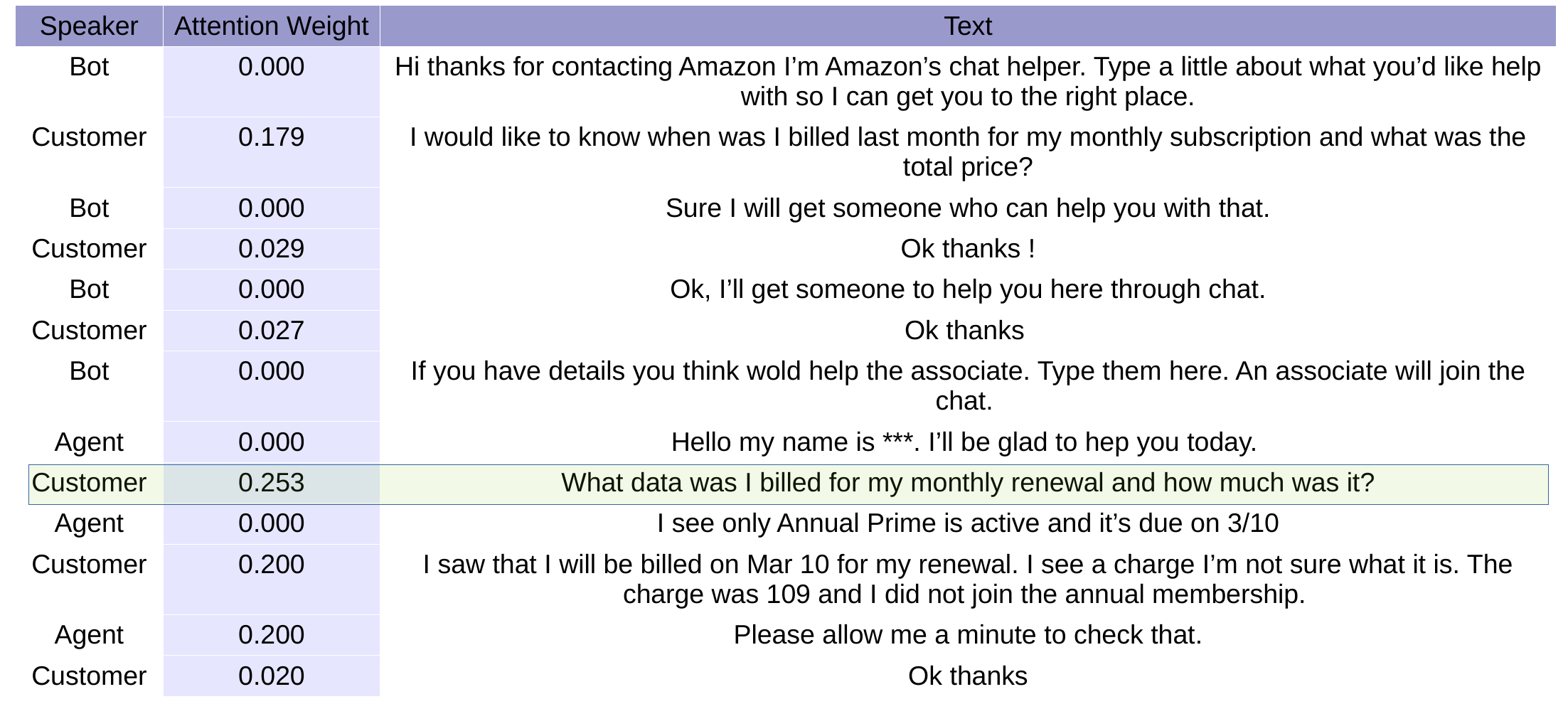}
  \caption{An illustration of the functioning of the SeaCat model (derived from a synthetic research dataset). The SeaCat model calculates the attention score, $\sigma_{i}$, for each sentence in a transcript. The sentence preceding and following the agent's first sentence, with the highest score is predicted as the customer's primary question, i.e. the highlighted sentence.
 }\label{seacat_example}
\end{figure*} 

\end{document}